\documentclass{article}

\usepackage[main, final]{neurips_2026}

\usepackage[utf8]{inputenc}
\usepackage[T1]{fontenc}
\usepackage{hyperref}
\usepackage{url}
\usepackage{booktabs}
\usepackage{amsmath}
\usepackage{multirow}
\usepackage{amsfonts}
\usepackage{nicefrac}
\usepackage{microtype}
\usepackage{xcolor}
\usepackage{graphicx}
\usepackage{float}
\usepackage[table]{xcolor}
\usepackage{wrapfig}
\usepackage{subcaption}
\usepackage{pifont}
\usepackage{algorithm}
\usepackage{algorithmic}
\usepackage{tabularx}
\newcolumntype{L}{>{\raggedright\arraybackslash}X}

\newcommand{\method}{\textsc{Teal}\xspace}
\definecolor{tealrow}{HTML}{E1F5EE}     %
\definecolor{tealink}{HTML}{085041}     %
\definecolor{gaingreen}{HTML}{1B7D3C}
\newcommand{\gain}[1]{\,{\scriptsize\textcolor{gaingreen}{(+#1)}}}

\usepackage{xspace}

\title{Pixels to Tokens: Token Space Efficient Active Learning  for Low-Budget Semantic Segmentation}

\author{%
  \textbf{Utku Cicek}$^{1}$ \quad \textbf{Mahip Singh}$^{1}$ \quad \textbf{Ziyao Shang}$^{1}$ \quad \textbf{Kimathi Kaai}$^{1,2}$ \\
  \textbf{C Thomas}$^{2}$ \quad \textbf{Pedro Pablo Guerrero Vela}$^{2}$ \quad \textbf{Alexander Wong}$^{1,2}$ \quad \textbf{Sirisha Rambhatla}$^{1}$ \\
  $^{1}$University of Waterloo \quad $^{2}$Apple \\
  \texttt{\{ucicek, mahip.singh, z27shang, kkaai, alexander.wong, sirisha.rambhatla\}@uwaterloo.ca} \\
  \texttt{\{c.thomas, pguerrerovela, alex\_wong3\}@apple.com}
}

\begin{document}

\maketitle

\begin{abstract}
While foundation models
and vision transformers (ViTs) have redefined visual representations, current active
learning (AL) strategies for these backbones largely operate at the patch or region
level, where annotation requirements remain high due to extensive pixel labeling. Extending ViT-based AL to the
pixel-level, low-budget regime is uniquely challenging: with as little as one pixel
query per image, a method must simultaneously identify the most informative
locations and train a reliable decoder head from an extremely sparse signal. We
introduce TEAL (Token-space Efficient Active Learning), the first pixel-level AL
framework built around ViT representations. TEAL uses a frozen ViT backbone
and performs diversity-based selection directly on its native token lattice, avoiding
the artifacts introduced by interpolating token embeddings to dense pixel resolutions.
Our framework applies a two-level MaxHerding strategy over multi-layer token
descriptors to select representative candidates, followed by margin-based uncertainty
refinement on a finer decoder grid. Across CamVid, Cityscapes, ADE20K, and
Pascal Context, TEAL consistently outperforms previous baselines under extreme
label scarcity, improving over the strongest baseline by up to $+\textbf{24.36}$ mIoU with a
single labeled pixel (px) per image, and reaching up to $\textbf{97\%}$ of fully supervised
performance with just 10 px. To rigorously establish the source of the gains, we further show that the improvements persist when all baselines are equipped with the same DINOv3 backbone, demonstrating that coupling token-space selection with uncertainty refinement on the decoder grid makes ViT token geometry an effective foundation for extremely low-budget active segmentation.
\end{abstract}
\section{Introduction}
\label{sec:Introduction}

\begin{figure}[t]
    \centering

    \begin{minipage}[t]{0.32\linewidth}
        \centering
        \includegraphics[width=\linewidth]{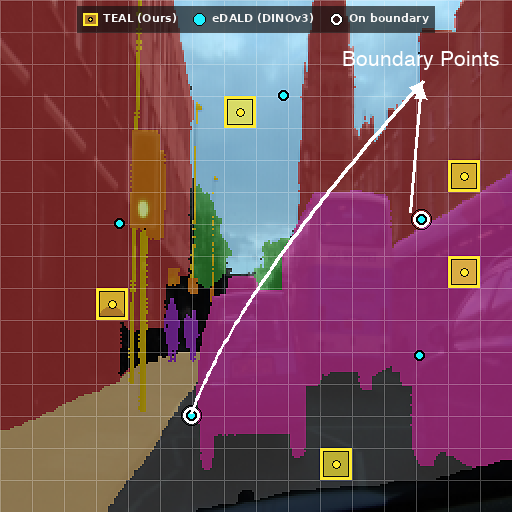}
        \vspace{2pt}
        {\small \textbf{(a) Query selection}}
    \end{minipage}
    \hfill
    \begin{minipage}[t]{0.32\linewidth}
        \centering
        \includegraphics[width=\linewidth]{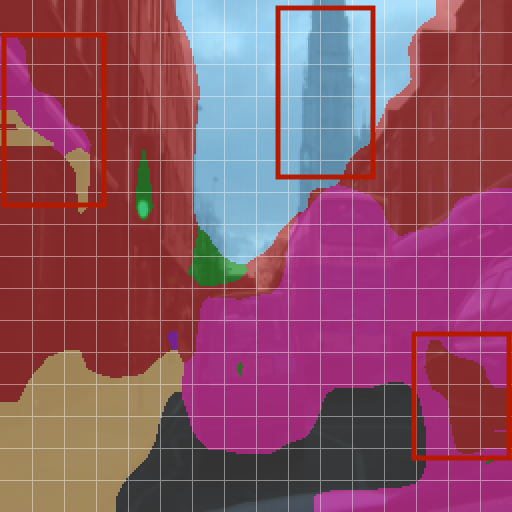}
        \vspace{2pt}
        {\small \textbf{(b) eDALD (DINOv3)}}
    \end{minipage}
    \hfill
    \begin{minipage}[t]{0.32\linewidth}
        \centering
        \includegraphics[width=\linewidth]{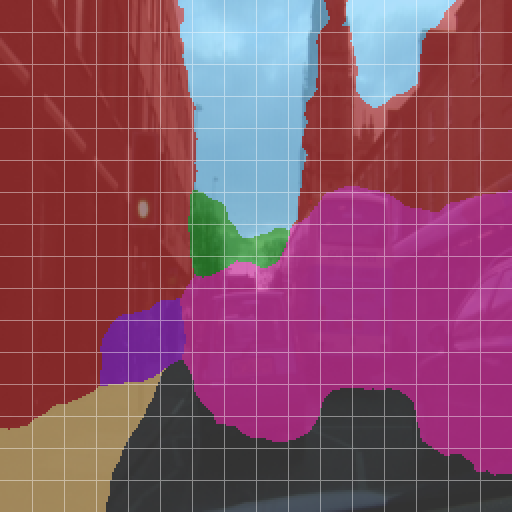}
        \vspace{2pt}
        {\small \textbf{(c) TEAL (Ours)}}
    \end{minipage}

    \caption{
    \textbf{Token-space acquisition avoids pixel-space interpolation artifacts.}
    (a) Compares query locations from TEAL and the two-stage eDALD baseline~\cite{kim2025diffusion}, with both methods using the same frozen DINOv3 backbone for a controlled comparison.
    Pixel-space selection often concentrates near boundaries, where upsampled token features mix neighboring descriptors, while TEAL selects more representative regions by operating on the native token lattice. As (c) shows that this leads to cleaner segmentation maps under the same sparse pixel-label budget.
    }
    \label{fig:teal_motivation}
\end{figure}

Semantic segmentation assigns a class label to every pixel in an image, but training accurate
segmentation models typically requires dense pixel annotations. Active learning approaches reduce
this cost by querying only a small set of informative pixels for annotation. Large-scale self-supervised pretraining has substantially improved dense visual representations, making ViT features a competitive backbone choice for semantic segmentation~\cite{dosovitskiy2021,caron2021emerging,oquab2023dinov2,simeoni2025dinov3}. Recent active learning methods have increasingly used vision transformer and foundation-model representations, showing that pretrained transformer features can serve as effective acquisition
spaces~\cite{gupte2024revisiting,mots2024gci,wu2025active}. However, these methods primarily
label images, patches, or regions rather than sparse individual pixels. Pixel-level active learning offers a fine control over labeling cost yet this setting is difficult because the acquisition space is extremely large,
highly redundant, and sensitive to the representation used for selection \cite{mittal2023best}.

Prior work has shown that segmentation models can learn from sparsely queried pixels~\cite{shin2021pixelpick}, and that low-budget pixel-level active learning benefits from combining diversity-based candidate selection with uncertainty-based refinement~\cite{kim2025diffusion}. However, these methods rely on CNN backbones whose lightweight per-pixel heads~\cite{chen2017rethinking,ho2020denoising} train reliably even under sparse supervision. ViT-based segmenters, by contrast, require heavier learned decoders (e.g., DPT) to project patch tokens to pixels — decoders that are themselves difficult to train under extreme budgets such as one annotated pixel per image, leaving it unclear whether reliable pixel-level active learning is achievable in this regime.

A natural approach is to upsample ViT token features to a dense pixel grid via interpolation, apply pixel-level acquisition directly on the resulting features, and train a lightweight pixel classifier. However, we find that this can lead to suboptimal annotation candidates for both uncertainty- and diversity-based selection. Both selectors tend to pick pixels near token boundaries because their descriptors are more affected by neighboring tokens. However, the embedding diversity and segmentation uncertainty of such pixels are due to interpolation, not genuine semantic ambiguity or visual diversity. As shown in Fig.~\ref{fig:teal_motivation}(a), pixel-space selection concentrates along grid-aligned boundaries, whereas token-space acquisition over native DINOv3 features lands on more representative interior regions, leading to better downstream segmentation (Fig.~\ref{fig:teal_motivation}(c) vs Fig.~\ref{fig:teal_motivation}(b)).

To this end, we introduce TEAL, a two-stage active learning framework for ViT-based pixel-level semantic
segmentation under extremely sparse budgets. TEAL progressively moves from token-level coverage
to pixel-level annotation. It first applies MaxHerding \cite{bae2024generalized} directly to ViT token features to select diverse
candidate tokens that cover the native representation space. It then refines these candidates using
uncertainty on a finer light-weight decoder grid and queries individual pixels for annotation. This token-to-pixel design preserves the semantic geometry of ViT features during acquisition while still producing
sparse pixel labels for training the downstream segmentation model.

Our contributions are threefold. First, we present, to our knowledge, the first pixel-level active learning framework built on ViT representations, showing that ViT-based segmentation can learn from extremely sparse pixel labels, contrary to prior findings that ViT features underperform diffusion features in this regime~\citep{kim2025diffusion}. With DINOv3~\citep{simeoni2025dinov3}, \textsc{Teal} outperforms the prior pixel-level baseline, eDALD with DDPM~\citep{ho2020denoising} features, by up to 24.36~mIoU across four datasets while being more efficient. Second, we show that these gains are not explained by the backbone alone: against matched-backbone DINOv3 baselines, \textsc{Teal} gains 0.50 to 2.47~mIoU at one pixel per image, widening up to 4~mIoU at $512 \times 512$ resolution. Third, our mechanistic analysis shows that pixel-space herding concentrates queries on token boundaries where interpolation creates artificial diversity, and that token-space coverage and cell-level refinement each contribute complementarily to closing this gap.

\section{Related Work}
\subsection{Semantic Segmentation}
Semantic segmentation assigns a class label to every pixel in an image. Early segmentation models were based on convolutional encoder-decoder architectures, such as U-Net~\cite{ronneberger2015}, SegNet~\cite{badrinarayanan2016}, and DeepLabv3~\cite{chen2017rethinking}. While these models remain strong dense prediction methods, their locality-biased convolutional representations can limit their ability to model long-range dependencies and global context. Vision Transformers (ViTs)~\cite{dosovitskiy2021} have significantly improved models' ability to recognize long-range dependencies, producing more capable and flexible architectures such as SegFormer~\cite{xie2021segformer} and Mask2Former~\cite{cheng2022masked}. More recently, large-scale self-supervised ViT backbones have become strong general purpose visual feature extractors. DINO shows that self-supervised ViT features can show both object level and segmentation relevant structure ~\cite{caron2021emerging}. DINOv2 improves transfer across dense visual tasks ~\cite{oquab2023dinov2}. DINOv3 further strengthens dense visual features using large-scale self-supervised training ~\cite{simeoni2025dinov3}. These models show that strong pretrained token representations can provide useful spatial and semantic structure even when supervision is limited.

Diffusion features have also been used for label-efficient semantic segmentation~\cite{baranchuk2022label} and recent low-budget active learning methods use diffusion features for candidate selection~\cite{kim2025diffusion}.
However, diffusion feature extraction is computationally expensive, and the acquisition is usually applied over dense pixel-space feature maps. In contrast, TEAL uses a frozen DINOv3 backbone and performs acquisition directly over its native token space.

\subsection{Active Learning}
Active learning (AL) reduces annotation cost by selecting the most informative data points to label. 
Recent work has also revisited AL in the era of vision foundation models, showing that strong pretrained representations affect key low-budget design choices like initial pool selection, diverse sampling and trade-off between representative and uncertainty-based acquisition ~\cite{gupte2024revisiting}. However, most foundation model AL studies focus on image-level classification rather than pixel-level semantic segmentation.
For semantic segmentation, the annotation unit is especially important. Image-level selection asks annotators to label full images, but can waste effort on redundant regions. Region-level methods use patches or superpixels, to reduce costs while preserving local context ~\cite{6248050, siddiqui2020viewpoint, golestaneh2020}. Pixel-level methods reduce the cost further by querying individual pixels.
A common family of methods selects samples using model uncertainty. ~\cite{siddiqui2020viewpoint} introduces \textit{viewpoint entropy}, leveraging model prediction inconsistencies under different views as an estimation of uncertainty. 
\textit{PixelPick}~\cite{shin2021pixelpick} shows that segmentation model can be trained from very sparse pixel labels using uncertainty-based pixel acquisition. Later methods, extend sparse pixel supervision with Bayesian or balanced uncertainty ~\cite{woo2023activ, didari2024}. 
These works establish pixel-level active learning as an effective low-budget setting, but their acquisition functions operate directly on dense pixel grids.

Another line of work is in diversity or coverage-based active learning. Instead of selecting only uncertain samples, coverage-based methods choose samples that represent the structure of images in the unlabeled pool.
Core-set selection uses feature space coverage for low-budget selection~\cite{sener2018active} while MaxHerding uses a kernel-based greedy objective for low-budget selection~\cite{bae2024generalized}.
Diffusion-driven two-stage active learning~\cite{kim2025diffusion} combines these ideas for semantic segmentation. It first uses MaxHerding over diffusion features to generate diverse pixel candidates, then refines them using uncertainty. 

However, this still performs candidate selection in dense pixel space. TEAL addresses this limitation. When the backbone is a ViT, pixel-space acquisition can misalign with pretrained patch token representation because upsampled pixel features are interpolated mixtures of patch token descriptors. Pixel-space coverage may select candidates that appear diverse because of boundary mixtures.

\section{Method}
Our method aligns active selection with the native token geometry of a frozen vision transformer.
Existing low-budget active learning approaches typically operate in a dense pixel space, both for
representation-based coverage and for uncertainty-based querying. However, when the underlying
backbone is a ViT, the representation is naturally defined over patch tokens rather than individual
pixels. Operating directly in pixel space can therefore introduce redundancy and distort the geometry
of the pretrained feature space. We instead propose a two-stage acquisition strategy that separates
\emph{semantic coverage} from \emph{spatial refinement}. Stage~1 performs diversity-based selection
directly in token space to identify a compact set of representative semantic regions, while Stage~2
refines these candidates using uncertainty on a finer decoder grid to choose the final pixel queries.
\subsection{Preliminaries}
\paragraph{Problem formulation.}
Let $\mathcal{D}=\{X_n\}_{n=1}^{N}$ be a collection of training images, and let $\Omega_n$ denote the set of pixel locations in image $X_n$. A frozen vision transformer backbone with patch size $p$ partitions each image into a grid of non-overlapping tokens $\mathcal{T}_n$, where each token $\tau \in \mathcal{T}_n$ corresponds to a $p \times p$ pixel region and is associated with a multi-scale feature vector $\mathbf{z}_\tau \in \mathbb{R}^d$, where d is embedding dimension.

At round $t$, we maintain a labeled pixel set $\mathcal{L}_t \subseteq \bigcup_{n=1}^{N}\Omega_n$ that is cumulated from all previous rounds, i.e., $\mathcal{L}_{t+1} = \mathcal{L}_{t} \cup
\mathcal{V}_{t}$, where $\mathcal{V}_{t}$ is the set of pixels selected in round $t$. Given a per-round annotation budget of $m$ pixels, our active learning pipeline proceeds as follows:
\begin{enumerate}
    \item \textbf{Diversity-based candidate selection.} Using the frozen token features $\{\mathbf{h}_\tau\}$, we apply a representation-based criterion to globally select a diverse candidate set of tokens $\mathcal{C}_t$ that broadly covers the feature space.
    
    \item \textbf{Uncertainty-based query selection.} A prediction head trained on the current labeled set $\mathcal{L}_t$ evaluates only the candidate tokens in $\mathcal{C}_t$ using an acquisition function. We then select the top-$m$ pixels scored by prediction uncertainty and add their ground-truth label to the labeled set, producing $\mathcal{L}_{t+1}$.
    
    \item \textbf{Model update.} The prediction head is retrained on the accumulated labels in $\mathcal{L}_{t+1}$, and the procedure repeats for the next round.
\end{enumerate}

\begin{figure}[t]
    \centering
    \includegraphics[width=1\linewidth]{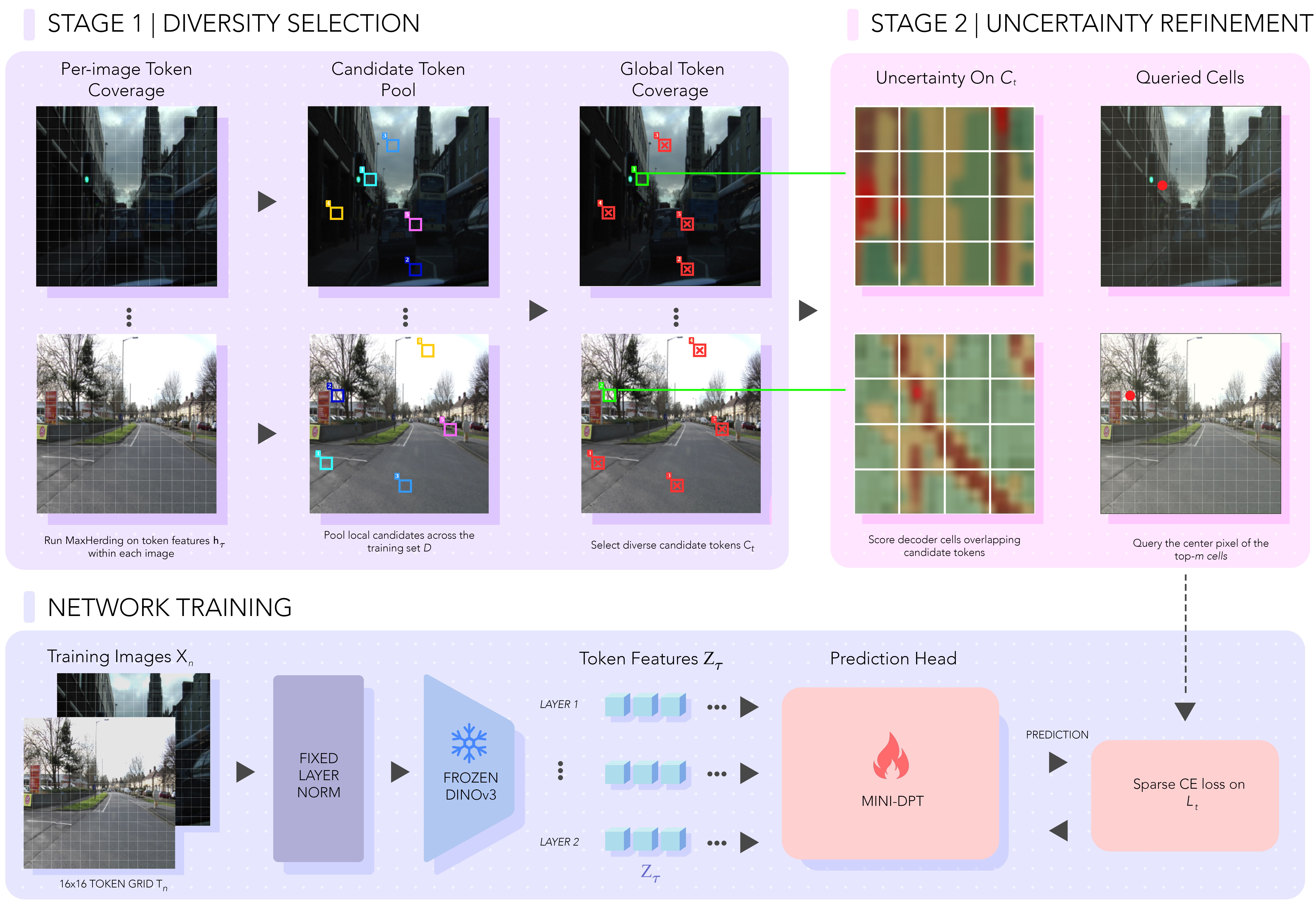}
    \caption{
\textbf{One round of TEAL active learning.} A Mini-DPT head is trained on the current labeled pixel set $\mathcal{L}_t$ using frozen multi-layer DINOv3 token features. Stage 1 performs two-level MaxHerding in the native token space: local herding selects $k$ representative unlabeled tokens per image, and global herding compresses their union into the dataset-level candidate set $\mathcal{C}_t$. Stage 2 refines these token candidates on the full $64 \times 64$ Mini-DPT decoder grid; in the figure, this is visualized locally by subdividing each selected candidate token into a $4 \times 4$ block of decoder cells. Margin uncertainty is evaluated only on valid decoder cells that overlap $\mathcal{C}_t$ and have unlabeled center pixels. The center pixels of the most uncertain valid cells are queried and added to $\mathcal{L}_{t+1}$ for the next round.
}
    \label{fig:met_figure}
\end{figure}

We focus on the extremely low-budget regime of 1 and 10 pixels per image on average. Our methodology is further outlined in Figure \ref{fig:met_figure}.

\paragraph{Segmentation head and training.}
Given an input image $X_n$, we use a frozen DINOv3-base backbone to extract token features from four transformer layers with indices $\{2,5,8,11\}$. For each token $\tau$, let $\mathbf{z}_{\tau}^{(\ell)} \in \mathbb{R}^{768}$ denote its embedding at layer $\ell$. We use multiple layers rather than a single layer so that the representation retains both lower-level visual detail from earlier layers and stronger semantic structure from deeper layers.

Because features from different transformer layers can have substantially different scales, we apply a fixed per-layer normalization before concatenation:
  \begin{equation}
  \hat{\mathbf{z}}_{\tau}^{(\ell)} =
  \frac{\mathbf{z}_{\tau}^{(\ell)}-\boldsymbol{\mu}^{(\ell)}}
       {\boldsymbol{\sigma}^{(\ell)}},
  \qquad
  \mathbf{z}_{\tau} = \left[\hat{\mathbf{z}}_{\tau}^{(2)};
  \hat{\mathbf{z}}_{\tau}^{(5)}; \hat{\mathbf{z}}_{\tau}^{(8)};
  \hat{\mathbf{z}}_{\tau}^{(11)}\right],
  \end{equation}
 where $\boldsymbol{\mu}^{(\ell)}$ and $\boldsymbol{\sigma}^{(\ell)}$ are per-layer mean and standard deviation, estimated once across all training-image tokens via Welford's online algorithm \cite{welford1962note} and frozen for the remainder of training.
The normalized features are then fed into an adapted \textbf{Mini-DPT} segmentation head~\cite{yang2025segdino}. We specifically chose Mini-DPT because it is well-suited to our sparse supervision setting. Its decoder reassembles multi-level token features and fuses them via residual convolutional blocks. This helps sparse supervision signals affect predictions not only at the labeled pixels themselves, but also at nearby pixels and visually similar regions with similar learned features. Since extremely sparse supervision increases the risk of overfitting in complex models, we further updated Mini-DPT to an even smaller model size. The adapted head predicts logits on a $64 \times 64$ grid (4x finer than the native $16 \times 16$ DINOv3 token grid). The logits are bilinearly upsampled to the original $256 \times 256$ image resolution.

At each active learning round, only the segmentation head is optimized, while the DINOv3 backbone remains frozen. The head is warm-started from the previous round and  trained for $E$ epochs on the accumulated labeled set $\mathcal{L}_t$ using a pixel-wise cross-entropy loss evaluated only at queried locations, with all unlabeled pixels assigned the ignore index.

\subsection{Coverage (Stage 1---Diversity Selection)}\label{sec:methodology-coverage}
In the extremely low-budget regime, selecting informative queries solely by uncertainty can easily concentrate the annotation budget on redundant regions, especially when many nearby locations share similar semantics. We therefore use a diversity-driven first stage to construct a candidate set $\mathcal{C}_t$ that broadly covers the feature space before uncertainty-based refinement.

Because DINOv3 is a vision transformer, its native representation is
defined on patch tokens rather than individual pixels. We therefore perform
Stage~1 in the token space to respect the geometry of the backbone
representation. For each token $\tau \in \mathcal{T}_n$, we build a herding
descriptor $\mathbf{h}_\tau \in \mathbb{R}^{d}$ in the same way as
$\mathbf{z}_\tau$, with fixed per-layer normalization and concatenation, but
from the final four layers $\{8,9,10,11\}$ rather than the head's layers.
Herding relies on distances in feature space, and these semantically mature
layers give more reliable distances.

  \paragraph{Coverage objective.}
  Token similarity is measured with an RBF kernel,
  \begin{equation}
  k(\mathbf{h}_i, \mathbf{h}_j) =
  \exp\!\left(
  -\frac{\|\mathbf{h}_i - \mathbf{h}_j\|_2^{\,2}}{\sigma^{2}}
  \right),
  \end{equation}
  where the bandwidth $\sigma$ is set to $s$ times the median pairwise
  distance between descriptors (we use $s = 2$), and the coverage of a
  subset $\mathcal{S} \subseteq \mathcal{P}$ with respect to a pool
  $\mathcal{P}$ is
  \begin{equation}
  \mathrm{Coverage}(\mathcal{S};\mathcal{P}) =
  \frac{1}{|\mathcal{P}|}
  \sum_{\tau \in \mathcal{P}}
  \max_{\tau' \in \mathcal{S}}
  k(\mathbf{h}_{\tau}, \mathbf{h}_{\tau'}).
  \end{equation}
    Given a pool $\mathcal{P}$ and a target \textit{size},
  $\mathrm{MaxHerd}(\mathcal{P}, size)$~\cite{bae2024generalized} starts
  from $\mathcal{S} = \emptyset$ and greedily adds the token that yields
  the largest increase in $\mathrm{Coverage}(\mathcal{S};\mathcal{P})$
  until $|\mathcal{S}| = size$.

  \paragraph{Two-level herding.}
  To balance per-image diversity with dataset-level coverage, we apply
  herding in two levels.  First, for each image $X_n$ we let
  $\mathcal{U}_{n,t} \subseteq \mathcal{T}_n$ denote the set of valid
  unlabeled tokens and select $k$ locally representative tokens per image,
  giving an aggregated pool
  \begin{equation}
  \mathcal{M}_t = \bigcup_{n=1}^{N}
  \mathrm{MaxHerd}\!\left(\mathcal{U}_{n,t}, size=k\right).
  \end{equation}

This prevents the candidate pool from collapsing onto a small number of images. Second, we apply herding globally on $\mathcal{M}_t$ to obtain the final candidate set $\mathcal{C}_t = \mathrm{MaxHerd}\!\left(\mathcal{M}_t,
size= |\mathcal{M}_t|/2\right)$ containing globally representative tokens. Each token in $\mathcal{C}_t$ corresponds to a $p \times p$ patch, from which the final set of pixels to be annotated will be selected in Stage 2 based on their uncertainty.

\subsection{Stage 2: Uncertainty Refinement}
Given the candidate token set $\mathcal{C}_t$ from Stage~1, Stage~2 selects the final pixels to annotate on a finer spatial grid. For each image $X_n$, the trained segmentation head outputs class logits on a $64 \times 64$ decoder grid before the final bilinear upsampling step. Each cell of this grid corresponds to a $4 \times 4$ image region.

 We score cells with margin sampling. For a cell $v$, the margin
  score is
  \begin{equation}
  u(v) = p_v^{(1)} - p_v^{(2)},
  \end{equation}
  where $p_v^{(1)}$ and $p_v^{(2)}$ are the top-two softmax class probabilities and smaller $u(v)$ corresponds to higher uncertainty.

From all training images, we select a subset of valid cells, where a cell is valid if and only if (i) its spatial support overlaps at least one candidate token in $\mathcal{C}_t$ and (ii) its center pixel is not already in $\mathcal{L}_t$. Uncertainty is evaluated only over this subset to prevent the acquisition from repeatedly focusing on previously annotated regions and to spread the budget across unexplored candidates.

All such valid cells are then pooled and ranked globally by uncertainty $u(v)$. For the $m$ cells with the smallest margin, we query their center pixel, forming the final selection for this round $\mathcal{V}_t$. Their ground-truth labels are added to $\mathcal{L}_{t+1} = \mathcal{L}_{t} \bigcup \mathcal{V}_t$. In this way, while Stage~1 ensures semantic diversity at the token level, Stage~2 refines the final query locations at the finer cell level.

\section{Experiments}
\subsection{Datasets}
We evaluate on four semantic segmentation benchmarks. CamVid \cite{brostow2009semantic} contains urban driving scenes captured in daytime and dusk conditions, originally at $960 \times 720$ resolution, with 11 classes. Cityscapes \cite{cordts2016cityscapes}  is a standard urban-scene benchmark with 2,975 training and 500 validation images across 19 classes. ADE20K \cite{zhou2017scene} contains diverse indoor and outdoor scenes with 964 training and 650 validation images. Pascal Context \cite{mottaghi2014role} provides dense scene-parsing labels; following prior work, we use the 33-class setting with 4,998 training and 5,105 validation images. All datasets are resized and center-cropped to $256 \times 256$; for ADE20K, we resize the shorter side to 256 while preserving aspect ratio before center-cropping following \cite{kim2025diffusion,gu2020context,mottaghi2014role}.
\subsection{Implementation details}
We evaluate our framework under two pixel-annotation budgets: 1 pixel and 10 pixels on average per image. As described in Section~\ref{sec:methodology-coverage}, we first apply MaxHerding~\cite{bae2024generalized} to the token set of each image and select $k = 20$ representative samples per image. These samples are then aggregated over the entire dataset to form an initial global pool, denoted by $M_t$. We subsequently apply MaxHerding once more to $M_t$ to construct the final candidate set $C_t$ which is half the size of initial global pool. The final candidate set of tokens, each covering a $16 \times 16$ pixel
patch, is then evaluated in Stage~2, where the center pixels of the top-$m$
most uncertain decoder cells are queried for labeling.
For each method, we train the segmentation head with their respective frozen backbones for 10 epochs with their own acquisition style, and we apply early stopping when training accuracy reaches 95\%. During training, we use the Adam  \cite{kingma2014adam} optimizer with a learning rate of $1\times10^{-4}$ and apply a pixel-level loss to the mini-DPT head \cite{yang2025segdino}. We then use validation-set performance as the fully supervised reference. All experiments are conducted
on NVIDIA RTX 6000 PRO GPUs.

\subsection{Main Results}
\label{sec:main-results}

\begingroup
\setlength{\intextsep}{0pt}

\begin{table}[t]
  \centering
  \small
  \caption{\textbf{Performance across datasets and labeling budgets}.
           Green values indicate absolute improvement over the strongest baseline.
           The \emph{Ceiling} column reports each method's fully supervised mIoU
           when trained with dense labels using its own backbone.
           PixelPick uses a CNN-based DeepLabV3 backbone
           \citep{chen2017rethinking}, 2-Stage eDALD uses DDPM
           \citep{ho2020denoising} features, and TEAL uses DINOv3-base
           \citep{simeoni2025dinov3} features with our token/grid-based
           segmentation head. Results are reported as mean $\pm$ standard deviation.}
  \label{tab:budget-results}

  \setlength{\tabcolsep}{10pt}
  \renewcommand{\arraystretch}{1.2}

  \resizebox{\textwidth}{!}{
  \begin{tabular}{@{}ll ll l@{}}
    \toprule
    \multirow{2}{*}{\textbf{Dataset}} &
    \multirow{2}{*}{\textbf{Method}} &
    \multicolumn{2}{c}{\textbf{Budget -- average pixels per image}} &
    \textbf{Ceiling} \\
    \cmidrule(lr){3-4}
    & & 1 & 10 & $256^2$ \\
    \midrule

    \multirow{3}{*}{CamVid}
      & PixelPick
      & 30.56 $\pm$ 1.58
      & 37.51 $\pm$ 0.66
      & 42.85 \\

      & 2-Stage eDALD
      & 34.15 $\pm$ 0.72
      & 40.20 $\pm$ 0.65
      & 58.27 \\

      & \cellcolor{tealrow}\textbf{\method}
      & \cellcolor{tealrow}%
        \textbf{49.54 $\pm$ 0.37}\textbf{\gain{15.39}}
      & \cellcolor{tealrow}%
        \textbf{55.93 $\pm$ 0.11}\textbf{\gain{15.73}}
      & \cellcolor{tealrow}62.02 \\

    \midrule

    \multirow{3}{*}{Cityscapes}
      & PixelPick
      & 25.96 $\pm$ 0.16
      & 31.03 $\pm$ 0.33
      & 32.38 \\

      & 2-Stage eDALD
      & 30.40 $\pm$ 0.41
      & 34.60 $\pm$ 0.47
      & 43.89 \\

      & \cellcolor{tealrow}\textbf{\method}
      & \cellcolor{tealrow}%
        \textbf{42.25 $\pm$ 0.37}\textbf{\gain{11.85}}
      & \cellcolor{tealrow}%
        \textbf{43.79 $\pm$ 0.26}\textbf{\gain{9.19}}
      & \cellcolor{tealrow}49.91 \\

    \midrule

    \multirow{3}{*}{ADE20K}
      & PixelPick
      & 11.06 $\pm$ 0.42
      & 16.35 $\pm$ 0.21
      & 18.89 \\

      & 2-Stage eDALD
      & 30.44 $\pm$ 0.40
      & 35.70 $\pm$ 1.16
      & 45.12 \\

      & \cellcolor{tealrow}\textbf{\method}
      & \cellcolor{tealrow}%
        \textbf{33.68 $\pm$ 0.46}\textbf{\gain{3.24}}
      & \cellcolor{tealrow}%
        \textbf{39.68 $\pm$ 0.60}\textbf{\gain{3.98}}
      & \cellcolor{tealrow}44.77 \\

    \midrule

    \multirow{3}{*}{Pascal Context}
      & PixelPick
      & 28.15 $\pm$ 0.88
      & 36.81 $\pm$ 0.55
      & 39.81 \\

      & 2-Stage eDALD
      & 48.23 $\pm$ 0.62
      & 58.77 $\pm$ 0.51
      & 65.34 \\

      & \cellcolor{tealrow}\textbf{\method}
      & \cellcolor{tealrow}%
        \textbf{72.59 $\pm$ 0.65}\textbf{\gain{24.36}}
      & \cellcolor{tealrow}%
        \textbf{75.30 $\pm$ 0.28}\textbf{\gain{16.53}}
      & \cellcolor{tealrow}77.53 \\

    \bottomrule
  \end{tabular}}
\end{table}

\begin{table}[t]
  \centering
  \small
  \caption{\textbf{Matched-backbone comparison across datasets and labeling budgets}.
           Green values indicate absolute improvement over the previous two-stage
           method \citep{kim2025diffusion}.
           \emph{Ceiling} reports the fully supervised mIoU obtained using dense
           labels and the DINOv3 backbone, shared by all methods.
           The final column reports the fraction of the fully supervised performance
           recovered at the 10-pixel budget.
           All methods use DINOv3 features for a matched-backbone comparison.}
  \label{tab:matched-backbone-results}

  \setlength{\tabcolsep}{10pt}
  \renewcommand{\arraystretch}{1.2}

  \resizebox{\textwidth}{!}{
  \begin{tabular}{@{}lc l ll c@{}}
    \toprule
    \multirow{2}{*}{\textbf{Dataset}} &
    \textbf{Ceiling} &
    \multirow{2}{*}{\textbf{Method}} &
    \multicolumn{2}{c}{\textbf{Budget -- average pixels per image}} &
    \textbf{\% of Sup.} \\
    \cmidrule(lr){4-5}
    & $256^2$ & & 1 & 10 & @10 px \\
    \midrule

    \multirow{3}{*}{CamVid}
      & \multirow{3}{*}{62.02}
      & PixelPick
      & 47.07 $\pm$ 0.28
      & 53.79 $\pm$ 0.27
      & 86.7 \\

      &
      & 2-Stage eDALD
      & 45.88 $\pm$ 1.68
      & 53.98 $\pm$ 0.57
      & 87.0 \\

      &
      & \cellcolor{tealrow}\textbf{\method}
      & \cellcolor{tealrow}%
        \textbf{49.54 $\pm$ 0.37}\textbf{\gain{3.66}}
      & \cellcolor{tealrow}%
        \textbf{55.93 $\pm$ 0.11}\textbf{\gain{1.95}}
      & \cellcolor{tealrow}\textbf{90.2} \\

    \midrule

    \multirow{3}{*}{Cityscapes}
      & \multirow{3}{*}{49.91}
      & PixelPick
      & 40.21 $\pm$ 0.27
      & 43.28 $\pm$ 0.17
      & 86.7 \\

      &
      & 2-Stage eDALD
      & 40.55 $\pm$ 0.41
      & 41.68 $\pm$ 0.20
      & 83.5 \\

      &
      & \cellcolor{tealrow}\textbf{\method}
      & \cellcolor{tealrow}%
        \textbf{42.25 $\pm$ 0.37}\textbf{\gain{1.70}}
      & \cellcolor{tealrow}%
        \textbf{43.79 $\pm$ 0.26}\textbf{\gain{2.11}}
      & \cellcolor{tealrow}\textbf{87.7} \\

    \midrule

    \multirow{3}{*}{ADE20K}
      & \multirow{3}{*}{44.77}
      & PixelPick
      & 30.88 $\pm$ 0.63
      & 36.67 $\pm$ 1.23
      & 81.9 \\

      &
      & 2-Stage eDALD
      & 31.69 $\pm$ 0.81
      & 38.66 $\pm$ 0.47
      & 86.4 \\

      &
      & \cellcolor{tealrow}\textbf{\method}
      & \cellcolor{tealrow}%
        \textbf{33.68 $\pm$ 0.46}\textbf{\gain{1.99}}
      & \cellcolor{tealrow}%
        \textbf{39.68 $\pm$ 0.60}\textbf{\gain{1.02}}
      & \cellcolor{tealrow}\textbf{88.6} \\

    \midrule

    \multirow{3}{*}{Pascal Context}
      & \multirow{3}{*}{77.53}
      & PixelPick
      & 70.92 $\pm$ 0.49
      & 74.76 $\pm$ 0.30
      & 96.4 \\

      &
      & 2-Stage eDALD
      & 72.09 $\pm$ 0.59
      & 74.49 $\pm$ 0.09
      & 96.1 \\

      &
      & \cellcolor{tealrow}\textbf{\method}
      & \cellcolor{tealrow}%
        \textbf{72.59 $\pm$ 0.65}\textbf{\gain{0.50}}
      & \cellcolor{tealrow}%
        \textbf{75.30 $\pm$ 0.28}\textbf{\gain{0.81}}
      & \cellcolor{tealrow}\textbf{97.1} \\

    \bottomrule
  \end{tabular}}
\end{table}

\endgroup

Table ~\ref{tab:budget-results} reports mIoU across four datasets under extremely sparse pixel-labeling budgets. TEAL consistently outperforms both PixelPick and 2-Stage eDALD across every dataset and budget, with gains over the strongest baseline ranging from +3.24 to +24.36 mIoU. Gains are largest on Pascal Context and CamVid, where large coherent semantic regions make DINOv3 tokens a strong acquisition space, and smaller on Cityscapes and ADE20K, where dense boundaries and fine structures make the low-resolution token-grid head the limiting factor. TEAL also recovers most of its fully supervised performance from only a handful of labels: at ten pixels per image, it reaches 90.2\% of its dense-label ceiling on CamVid, 87.7\% on Cityscapes, 88.6\% on ADE20K, and 97.1\% on Pascal Context.

Table ~\ref{tab:matched-backbone-results} tests whether these gains merely reflect the backbone by equipping all methods with DINOv3 features. The baselines improve substantially, but TEAL remains best on every dataset and budget. The same holds relative to the ceiling: at ten pixels per image, TEAL recovers the largest fraction of fully supervised performance on all four datasets, exceeding the pixel space baseline \citep{kim2025diffusion} by up to 3.2 points of the ceiling on CamVid. With the representation held fixed, the acquisition strategy alone determines how close each method gets to dense-label performance, and TEAL's token-space selection closes more of that gap. Overall, these results support our claim that DINOv3-based vision transformers are well suited for low-budget pixel-level active segmentation, and that TEAL's improvements come from both the representation and the acquisition mechanism.

\section{Discussion}
\label{sec:Discussion}
To isolate the contribution of the acquisition procedure from backbone
capacity, we run mechanistic analyses to understand why token-space
acquisition outperforms its pixel-space counterpart and how the two stages
specialize across budgets. As the pixel-space counterpart, we use
\emph{pixel herding}: MaxHerding run on DINOv3 features bilinearly upsampled
to pixel resolution, with pixels then chosen directly by margin uncertainty,
which is equivalent to matched-backbone eDALD with margin sampling in place
of eBALD \cite{kim2025diffusion}. Unlike TEAL, pixel herding computes coverage over pixels rather than native
tokens, so its uncertainty stage can only rank the individual pixels that
herding has already selected. TEAL instead selects whole tokens, so its
uncertainty stage can choose among all decoder cells within each selected
token's $16 \times 16$ pixel patch. We also
test how sensitive TEAL is to its design choices. All analyses use the same
DINOv3 backbone. Every variant uses margin sampling as its acquisition
function, except eDALD DINO, which keeps its original eBALD criterion as a
reference point.

\begin{figure}[t]
    \centering
    \includegraphics[width=\linewidth]{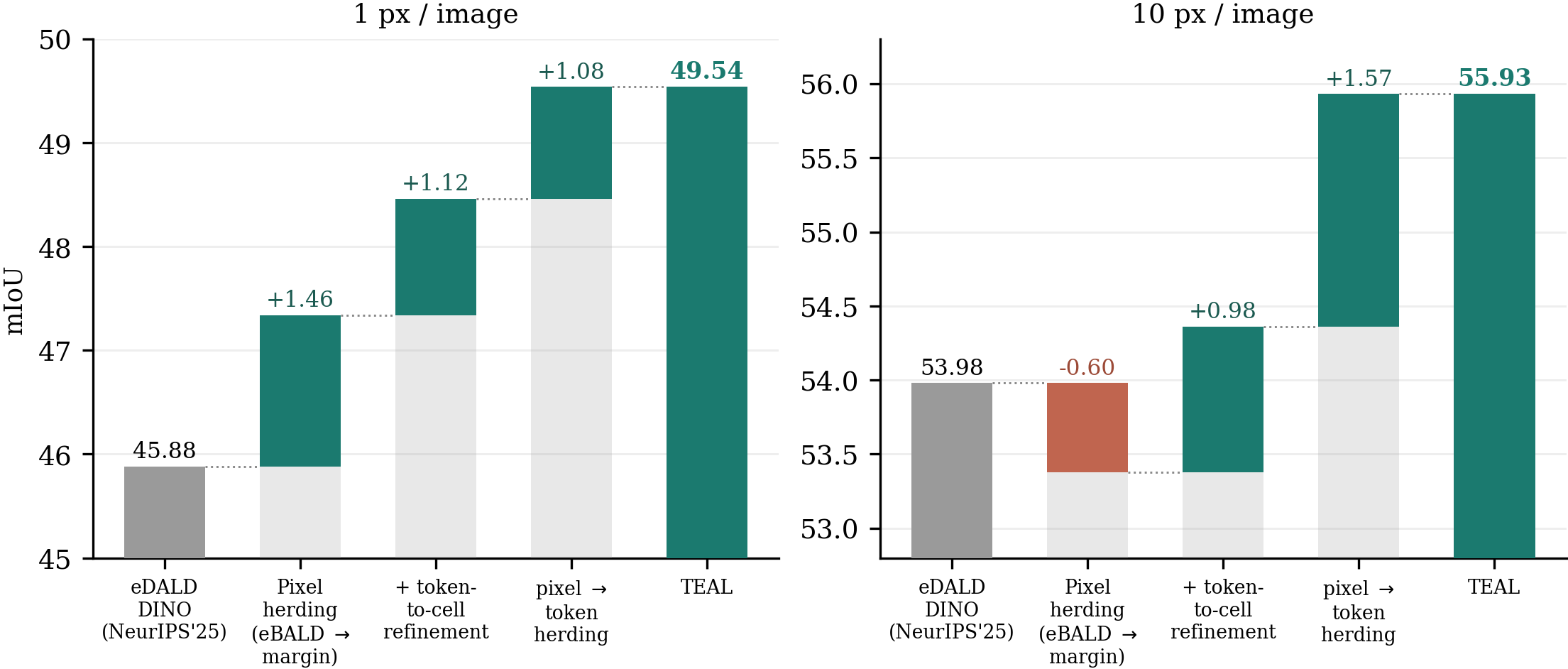}
    \caption{\textbf{Decomposing the gap from matched-backbone eDALD to
    \method{} on CamVid.} Solid segments show the change from the previous
    configuration ($\rightarrow$ replaces, $+$ adds; red marks a loss). Pixel
    herding helps at one pixel per image but hurts at ten, while token-to-cell
    refinement and token-space herding help at both.}
    \label{fig:component-stacked}
\end{figure}

\paragraph{Pixel-space herding queries interpolation artifacts.}
Section ~\ref{sec:Introduction} hypothesized that upsampling token features
creates artificial diversity near token boundaries. We test this by
measuring where queries land relative to the token grid: $60.1\%$ of
pixel-herding queries fall exactly on token edges, versus $0\%$ for
token-space herding and $23.4\%$ for random selection, more than twice the
rate chance would produce (Table \ref{tab:token_edge}). Figure~\ref{fig:component-stacked} attributes the
resulting $2.2$ mIoU gap between pixel herding
($47.34$) and TEAL ($49.54$) on CamVid at one pixel per image: token-to-cell
refinement recovers $1.12$ mIoU by moving queries off the artifact, and the
remaining $1.08$ is closed only by selecting candidates in the native token
space. The same pattern holds at ten pixels per image, where the gap widens
to $2.55$ mIoU ($53.38$ to $55.93$): refinement contributes $0.98$ and
token-space coverage the remaining $1.57$. Both stages therefore contribute
at both budgets, and operating on the token lattice removes the artificial
diversity at its source.
\paragraph{The two stages specialize by budget.}
At one pixel per image, TEAL prioritizes spread: only $2\%$ of its labeled tokens receive more than one label, and its rare-class share exceeds
matched-backbone eDALD's ($14.5\%$ vs.\ $10.1\%$). At ten pixels per image, TEAL instead resolves region boundaries by
allocating more than one pixel to $29.8\%$ of the tokens it selects, versus
$1.8\%$ for eDALD, while still keeping a higher rare-class share
($12.7\%$ vs.\ $7.5\%$, Table~\ref{tab:appendix-query-allocation}). Lacking this refinement, eDALD becomes less competitive as the budget
grows, falling behind the purely uncertainty-based PixelPick at ten pixels
per image on Cityscapes and Pascal Context.

\begin{figure}[t]
    \centering
    \includegraphics[width=\linewidth]{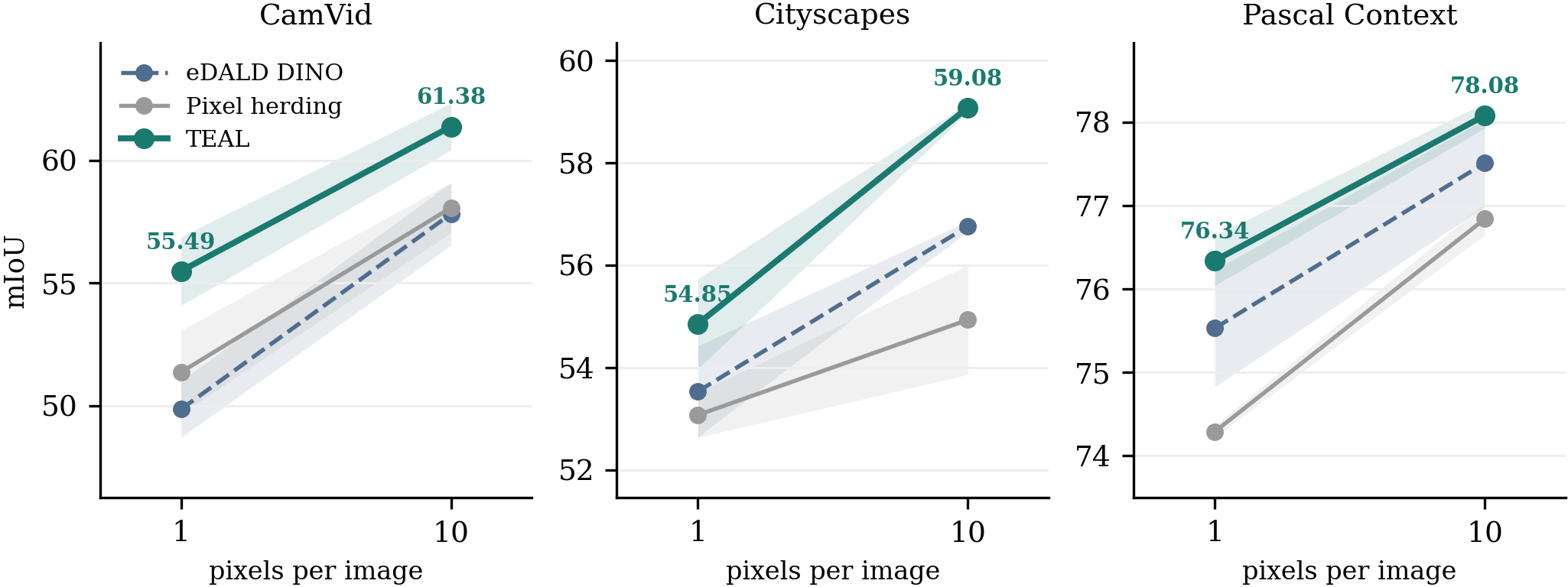}
    \caption{\textbf{Results at $512\times512$ resolution.} Matched-backbone
    comparison of eDALD DINO, pixel herding (eDALD DINO with margin sampling),
    and \method{} at 1 and 10 pixels per image. Shaded bands show the standard
    deviation over three seeds.}
    \label{fig:512_resolution}
\end{figure}

\paragraph{Resolution scaling.}
At $512\times512$ on Cityscapes, a spatially complex benchmark with dense
boundaries, thin structures, and small objects, TEAL reaches
$54.85 \pm 0.88$ mIoU at one pixel per image and $59.08 \pm 0.09$ at ten.
This improves over the strongest baseline, eDALD DINO ($53.53 \pm 0.89$ and
$56.76 \pm 0.10$), by $+1.32$ and $+2.32$ mIoU, and over pixel herding
($53.07 \pm 0.44$ and $54.93 \pm 1.07$) by $+1.78$ and $+4.15$. This
observation extends to the other datasets, as shown in
Figure~\ref{fig:512_resolution}: TEAL improves over eDALD DINO by $+5.63$
and $+3.56$ mIoU on CamVid, and by $+0.81$ and $+0.57$ on Pascal Context, at
one and ten pixels per image respectively.

\paragraph{Sensitivity and design choices.}
TEAL is robust to its main hyperparameters: performance varies by under
$1.2$ mIoU across local budgets $k \in [5,10,20,40,80]$ and RBF bandwidths
$\sigma \in [0.5, 3.0]$, degrading only when the candidate pool becomes too
small ($K \in 5$) to capture token diversity. Layer selection matters more: herding on
the final consecutive DINOv3 layers $[8,9,10,11]$ outperforms shallow
layers ($30.30$ $\pm 3.85$), and balanced layers $[2,5,8,11]$ ($48.80$ $\pm 0.58$). Both late
layers ($50.21$ $\pm 1.29$) and deep layers ($49.54$ $\pm 0.37$) perform the best as semantically mature late-layer
features provide more reliable distances for coverage (Table \ref{tab:sens-layers}). We use the deep configuration due to its
low variance. 

\paragraph{Normalization.}
Fixed feature normalization also contributes to performance. Without fixed normalization, performance drops to $47.90$ $\pm$ 0.36 mIoU, and replacing it with a learned normalization layer only slightly improves the performance to $48.20$ $\pm$ 1.21 mIoU, compared to 49.54 $\pm$ 0.37 mIoU with fixed normalization.  We hypothesize that learned normalization is difficult to estimate reliably in the low-budget regime, where only a small number of labels are available in each round. As a result, its parameters may overfit and vary across rounds, making the feature space unstable for both selection and training. In contrast, fixed normalization preserves a consistent feature scale, which appears to be much more reliable under sparse supervision.

\paragraph{Acquisition function.}
    Our frozen DINOv3 encoder is deterministic, so BALD-style criteria such as
    eDALD's \citep{kim2025diffusion} eBALD require Monte Carlo (MC) dropout in the head, whose disagreement under
    sparse supervision mostly reflects an undertrained head rather than genuine
    uncertainty. Replacing margin with BALD or eBALD inside TEAL drops CamVid
    performance at one pixel per image from $49.54$ $\pm 0.37$ to $44.94$ $\pm 0.93$ and $45.02$ $\pm 1.79$
    mIoU. The gap narrows at ten pixels per image ($55.38$ $\pm 0.51$ and $55.51$ $\pm 0.04$ vs.
    $55.93$ $\pm 0.11$), as more labels stabilize the head, making margin the more reliable
    choice under extreme sparsity.

\paragraph{Backbone ablation.} 
On CamVid at one pixel per image, performance within the same TEAL pipeline improves from GoogleViT ($43.78 \pm 0.56$) to DINOv2 ($47.58 \pm 0.87$) to DINOv3 ($49.54 \pm 0.37$). Since herding relies on distances in feature space, it benefits when nearby
semantic regions are represented more consistently and when object boundaries are preserved more
clearly. In this setting, better local structure makes token embeddings more reliable as diversity units. Even the weakest ViT backbone outperforms the
CNN-based PixelPick ($30.56$) and diffusion-based eDALD ($34.15$) from
Table~\ref{tab:budget-results}, showing that ViT features as a whole are
well suited to the low-budget regime.

\paragraph{Limitations}
Although the runs remain reasonably fast in our setting, as reported in
Table~\ref{tab:efficiency}, TEAL becomes more expensive as image resolution increases. The main
bottleneck is the DINO forward pass: higher-resolution inputs produce more ViT tokens, and
self-attention becomes increasingly costly as the token count grows. Higher resolution images lead to better segmentation results, but this can also slow acquisition and
training-time feature extraction. A second limitation is the lightweight Mini-DPT
decoder. Its low-capacity design is useful under sparse supervision because it reduces overfitting and
allows the model to learn from very few labeled pixels. However, the same design can struggle on
complex scenes with many object boundaries, where stronger spatial modeling and higher-resolution
decoding are needed for precise segmentation.

\section{Conclusion}
We presented \method, a low-budget active learning framework for semantic segmentation that performs two-stage diversity-then-uncertainty acquisition in the native DINOv3 token space and trains a lightweight Mini-DPT head from sparse pixel labels. Across CamVid, Cityscapes, ADE20K, and Pascal Context, \method\ consistently improves over strong low-budget baselines and recovers a large fraction of fully supervised performance from only a handful of labels. Matched-backbone comparisons further indicate that these gains stem from two complementary choices: token-space acquisition and cell-level uncertainty refinement. More broadly, our results suggest that active learning for segmentation benefits from respecting the structure of self-supervised transformer features. A natural next step is to exploit DINOv3's similarity structure beyond acquisition: because semantically related regions stay close in token space, label propagation could densify supervision across nearby and cross-image tokens without additional annotation.

\section{Acknowledgment}
We would like to acknowledge the support of the Natural Sciences and Engineering Research Council of Canada (NSERC) Alliance Grant ALLRP 587270-2023.

\clearpage
\bibliographystyle{plainnat}
\bibliography{references}

\clearpage
\appendix
\section{Hyperparameters}

\subsection{Hyperparameter sensitivity}
\label{app:sensitivity}

We vary one hyperparameter at a time on CamVid at 1 pixel per image,
keeping all others at their defaults (shaded in each table). All sweeps
report the mean $\pm$ standard deviation over three seeds.

\subsubsection{Local budget $k$}

$K$ sets how many representative tokens local herding selects from each
image before global herding. Performance peaks at the default $k=20$
($49.54$ mIoU). Smallest pools hurt the most ($46.15$ at $k=5$) 
since too few tokens per image cannot cover that
image's variation. Larger pools degrade more gently
($48.40$ at $k=80$), likely because more redundant tokens enter the global
pool.

\begin{table}[H]
\centering
\small
\caption{Sensitivity to the local budget $k$.}
\label{tab:sens-local-budget}
\begin{tabular}{@{}lc@{}}
\toprule
\textbf{Local budget $k$} & \textbf{mIoU} \\
\midrule
5  & 46.15 $\pm 1.53$ \\
10 & 49.02 $\pm 2.10$ \\
\rowcolor{tealrow}
20 (default) & 49.54 $\pm 0.37$ \\
40 & 49.43 $\pm 0.43$ \\
80 & 48.40 $\pm 0.75$ \\
\bottomrule
\end{tabular}
\end{table}

\subsubsection{Global retention ratio}

The retention ratio sets what fraction of the locally selected tokens global
herding keeps as the final candidate set. Performance is stable between
$0.25$ and $0.75$ ($49.29$ to $50.43$ mIoU). Keeping every candidate
($1.00$) drops performance to $48.80$ $\pm 1.21$: without global compression, similar
tokens from different images stay in the candidate set, and Stage~2 can
spend the budget on redundant regions. We use $0.50$ because it halves the
candidate set, reducing Stage~2 cost, while staying within $0.9$ mIoU of the
best setting.

\begin{table}[H]
\centering
\small
\caption{Sensitivity to the global retention ratio.}
\label{tab:sens-global-retention}
\begin{tabular}{@{}lc@{}}
\toprule
\textbf{Global retention ratio} & \textbf{mIoU} \\
\midrule
0.25 & 49.29 $\pm 0.78$ \\
\rowcolor{tealrow}
0.50 (default) & 49.54 $\pm 0.37$ \\
0.75 & 50.43 $\pm 0.11$ \\
1.00 & 48.80 $\pm 1.21$ \\
\bottomrule
\end{tabular}
\end{table}

\subsubsection{RBF bandwidth scale factor $s$}

The RBF bandwidth is not set to an absolute value but scaled from the data:
$\sigma = s$ times the median pairwise distance between token descriptors.
Absolute bandwidths in a range like $[0.5, 3.0]$ are far smaller than the
typical distance between these features and drive the kernel to zero, so
scaling by the median keeps the kernel in a meaningful range regardless of
feature magnitude. TEAL is generally robust to its scaling factor, except when the scale factor becomes too small ($s$=1.0) treating most tokens as distinct. 

\begin{table}[H]
\centering
\small
\caption{Sensitivity to the RBF bandwidth scale factor $s$, where
$\sigma = s \cdot$ median pairwise token distance.}
\label{tab:sens-rbf}
\begin{tabular}{@{}lc@{}}
\toprule
\textbf{Scale factor $s$} & \textbf{mIoU} \\
\midrule
1.0 & 48.48 $\pm 0.78$ \\
\rowcolor{tealrow}
2.0 (default) & 49.54 $\pm 0.37$ \\
3.0 & 49.45 $\pm 0.95$ \\
\bottomrule
\end{tabular}
\end{table}

\subsubsection{Herding layers}

This sweep changes which DINOv3 layers provide the features used for
herding; the segmentation head keeps layers $[2,5,8,11]$ throughout. Layer
choice has the largest effect of any hyperparameter. Shallow layers perform
poorly ($30.30$ mIoU), since early features mostly encode low-level
appearance, so coverage over them spreads queries across texture rather than
semantic variation. Performance improves with the inclusion of more mature layers: $48.80$ $\pm 0.58$
for the Mini-DPT default, $50.21$ $\pm 1.29$  for the late configuration, and $49.54$ $\pm 0.37$ for
the final four consecutive layers. We use the deep configuration due to its low variance. 

\begin{table}[H]
\centering
\small
\caption{Sensitivity to the DINOv3 layers used for herding.}
\label{tab:sens-layers}
\begin{tabular}{@{}lc@{}}
\toprule
\textbf{Herding layers} & \textbf{mIoU} \\
\midrule
Shallow $[0,1,2,3]$ & 30.30 $\pm 3.85$ \\
Balanced - Mini-DPT default $[2,5,8,11]$~\citep{yang2025segdino} & 48.80 $\pm 0.58$ \\
Late $[5,7,9,11]$ & 50.21 $\pm 1.29$\\
\rowcolor{tealrow}
Deep $[8,9,10,11]$ (ours) & 49.54  $\pm 0.37$ \\
\bottomrule
\end{tabular}
\end{table}

\section{Additional Results}

\subsection{Pixel herding across datasets}
\label{app:pixel-herding}

Table~\ref{tab:appendix-pixel-herding} extends the pixel herding comparison
from Section~\ref{sec:Discussion} to all four datasets. Pixel herding shares
TEAL's backbone, head, and margin acquisition function, but computes
coverage over upsampled pixel features and then ranks only the pixels that
herding has already selected by uncertainty, whereas TEAL computes
uncertainty over all decoder cells within each selected token. TEAL
outperforms pixel herding on every dataset and budget, by $1.19$ to $2.55$
mIoU at one pixel per image and $1.31$ to $2.55$ at ten.

\begin{table}[H]
\centering
\caption{\textbf{Pixel herding vs.\ \method{} under matched backbones.}
mIoU at 1 and 10 pixels per image, mean $\pm$ std over three seeds. Green
values indicate absolute improvement over pixel herding.}
\label{tab:appendix-pixel-herding}

\vspace{0.4em}

\scriptsize
\setlength{\tabcolsep}{6pt}
\renewcommand{\arraystretch}{1.1}
\begin{tabular}{@{}ll cc@{}}
\toprule
\multirow{2}{*}{\textbf{Dataset}} &
\multirow{2}{*}{\textbf{Method}} &
\multicolumn{2}{c}{\textbf{Budget -- average pixels per image}} \\
\cmidrule(lr){3-4}
& & 1 & 10 \\
\midrule

\multirow{2}{*}{CamVid}
  & Pixel herding
  & 47.34 $\pm$ 0.29
  & 53.38 $\pm$ 0.50 \\
  & \cellcolor{tealrow}\textbf{\method}
  & \cellcolor{tealrow}\textbf{49.54 $\pm$ 0.37}\textbf{\gain{2.20}}
  & \cellcolor{tealrow}\textbf{55.93 $\pm$ 0.11}\textbf{\gain{2.55}} \\

\midrule

\multirow{2}{*}{Cityscapes}
  & Pixel herding
  & 40.61 $\pm$ 0.34
  & 41.80 $\pm$ 0.31 \\
  & \cellcolor{tealrow}\textbf{\method}
  & \cellcolor{tealrow}\textbf{42.25 $\pm$ 0.37}\textbf{\gain{1.64}}
  & \cellcolor{tealrow}\textbf{43.79 $\pm$ 0.26}\textbf{\gain{1.99}} \\

\midrule

\multirow{2}{*}{ADE20K}
  & Pixel herding
  & 32.67 $\pm$ 0.39
  & 38.31 $\pm$ 0.54 \\
  & \cellcolor{tealrow}\textbf{\method}
  & \cellcolor{tealrow}\textbf{33.68 $\pm$ 0.46}\textbf{\gain{1.01}}
  & \cellcolor{tealrow}\textbf{39.68 $\pm$ 0.60}\textbf{\gain{1.37}} \\

\midrule

\multirow{2}{*}{Pascal Context}
  & Pixel herding
  & 71.32 $\pm$ 0.22
  & 74.11 $\pm$ 0.32 \\
  & \cellcolor{tealrow}\textbf{\method}
  & \cellcolor{tealrow}\textbf{72.59 $\pm$ 0.65}\textbf{\gain{1.27}}
  & \cellcolor{tealrow}\textbf{75.30 $\pm$ 0.28}\textbf{\gain{1.19}} \\

\bottomrule
\end{tabular}
\end{table}
\label{app:512}

\subsection{Results at higher resolution}
\begin{table}[H]
\centering
\caption{\textbf{Results at $512\times512$ resolution.} Matched-backbone
comparison at 1 and 10 pixels per image, mean $\pm$ std over three seeds.
Pixel herding is eDALD DINO with margin sampling in place of eBALD.}
\label{tab:appendix-512}

\vspace{0.4em}

\scriptsize
\setlength{\tabcolsep}{6pt}
\renewcommand{\arraystretch}{1.1}
\begin{tabular}{@{}ll cc@{}}
\toprule
\multirow{2}{*}{\textbf{Dataset}} &
\multirow{2}{*}{\textbf{Method}} &
\multicolumn{2}{c}{\textbf{Budget -- average pixels per image}} \\
\cmidrule(lr){3-4}
& & 1 & 10 \\
\midrule

\multirow{3}{*}{CamVid}
  & eDALD DINO
  & 49.86 $\pm$ 1.15
  & 57.82 $\pm$ 1.28 \\
  & Pixel herding
  & 51.37 $\pm$ 1.70
  & 58.07 $\pm$ 1.02 \\
  & \cellcolor{tealrow}\textbf{\method}
  & \cellcolor{tealrow}\textbf{55.49 $\pm$ 1.40}
  & \cellcolor{tealrow}\textbf{61.38 $\pm$ 0.96} \\

\midrule

\multirow{3}{*}{Cityscapes}
  & eDALD DINO
  & 53.53 $\pm$ 0.89
  & 56.76 $\pm$ 0.10 \\
  & Pixel herding
  & 53.07 $\pm$ 0.44
  & 54.93 $\pm$ 1.07 \\
  & \cellcolor{tealrow}\textbf{\method}
  & \cellcolor{tealrow}\textbf{54.85 $\pm$ 0.88}
  & \cellcolor{tealrow}\textbf{59.08 $\pm$ 0.09} \\

\midrule

\multirow{3}{*}{Pascal Context}
  & eDALD DINO
  & 75.53 $\pm$ 0.70
  & 77.51 $\pm$ 0.52 \\
  & Pixel herding
  & 74.29 $\pm$ 0.07
  & 76.84 $\pm$ 0.19 \\
  & \cellcolor{tealrow}\textbf{\method}
  & \cellcolor{tealrow}\textbf{76.34 $\pm$ 0.30}
  & \cellcolor{tealrow}\textbf{78.08 $\pm$ 0.15} \\

\bottomrule
\end{tabular}
\end{table}

To test whether TEAL's advantage holds when the token grid is finer, we
repeat the matched-backbone comparison at $512\times512$, where each image
yields four times as many DINOv3 tokens as at $256\times256$
(Table~\ref{tab:appendix-512}). TEAL outperforms both eDALD DINO and pixel
herding on all three datasets at both budgets. Against eDALD DINO, the gain
is $+5.63$ mIoU at one pixel per image and $+3.56$ at ten on CamVid, and
$+0.81$ and $+0.57$ on Pascal Context. On Cityscapes, the dataset with the
densest boundaries and smallest structures, the gain grows with the budget,
from $+1.32$ to $+2.32$, suggesting that the finer token grid gives TEAL's
cell-level refinement more room to resolve boundaries.

\subsection{Query allocation}
\label{app:query-allocation}

\begin{table}[H]
\centering
\caption{\textbf{Fraction of selected pixels lying on or near token
boundaries.}}
\label{tab:token_edge}

\vspace{0.4em}

\scriptsize
\setlength{\tabcolsep}{6pt}
\renewcommand{\arraystretch}{1.1}
\begin{tabular}{@{}l ccc@{}}
\toprule
\textbf{Method} &
\textbf{On token edge (0\,px)} &
\textbf{Within 1\,px} &
\textbf{Within 2\,px} \\
\midrule
Pixel herding  & 60.1\% & 87.8\% & 97.0\% \\
\rowcolor{tealrow}
\method        &  0.0\% & 40.7\% & 69.6\% \\
Uniform random & 23.4\% & 43.8\% & 60.9\% \\
\bottomrule
\end{tabular}
\end{table}
\begin{table}[H]
\centering
\caption{\textbf{Query allocation under matched backbones on CamVid.}
Rare-class share is the fraction of queries placed on rare classes;
multi-label tokens is the fraction of labeled tokens that received more than one label.}
\label{tab:appendix-query-allocation}

\vspace{0.4em}

\scriptsize
\setlength{\tabcolsep}{6pt}
\renewcommand{\arraystretch}{1.1}
\begin{tabular}{@{}l ccc@{}}
\toprule
\textbf{Label set (matched backbone)} &
\textbf{Rare-class share} &
\textbf{Multi-label tokens} &
\textbf{Final mIoU} \\
\midrule
\rowcolor{tealrow}
\method, 1 px      & 14.5\% & 2.0\%  & 49.54 $\pm$ 0.37 \\
eDALD DINO, 1 px   & 10.1\% & 0.0\%  & 45.88 $\pm$ 1.68 \\
\midrule
\rowcolor{tealrow}
\method, 10 px     & 12.7\% & 29.8\% & 55.93 $\pm$ 0.11 \\
eDALD DINO, 10 px  & 7.5\%  & 1.8\%  & 53.98 $\pm$ 0.57 \\
\bottomrule
\end{tabular}
\end{table}

\begin{figure}[H]
    \centering
    \includegraphics[width=1\linewidth]{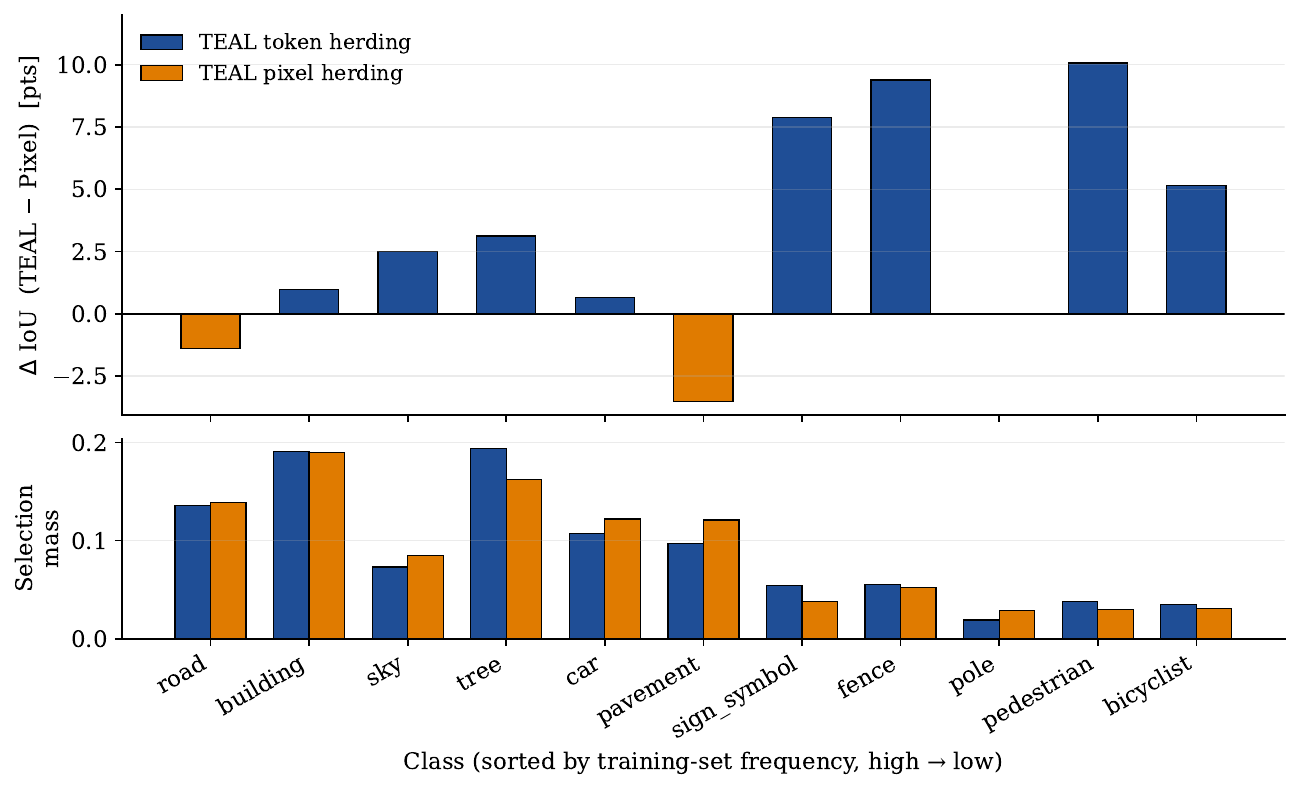}
    \caption{\textbf{Per-class IoU and selection mass at 10 pixels per image.}
    Comparison of token-level and pixel-level herding using DINOv3 features on
    CamVid. The top panel reports the per-class IoU difference between token
    and pixel herding (positive values favor token herding), and the bottom
    panel shows the fraction of selected queries assigned to each class.
    Token herding improves IoU for most classes, with the largest gains on
    lower-frequency, spatially localized categories such as sign symbol,
    fence, pedestrian, and bicyclist. Pixel herding is slightly better only on
    large surface classes such as road and pavement.}
    \label{fig:Selection_mass}
\end{figure}
\label{app:details}

\subsection{Compute Cost}
 \begin{table}[H]
    \centering
    \small
    \setlength{\tabcolsep}{4pt}
        \caption{%
      \textbf{Compute cost of CamVid run across different methods.}
      Per-image latency for Stage~1 herding,
      head training step, Stage~2 uncertainty, and the total
      wall-clock for a full 10-round run.
    }
    \label{tab:efficiency}
    \begin{tabular}{l rrr r}
      \toprule
                       & \multicolumn{3}{c}{Per-image latency (ms)}                 & Wall-Clock Runtime       \\
      \cmidrule(lr){2-4}
      Method           & Stage~1                  & Train step                 & Stage~2                 & 10-round AL run        \\
      \midrule
      eDALD-DDPM       & $184.3$                & $112.7$               & $555.7$              & $1\text{h}\,33\text{min}$ \\
      \textbf{TEAL (ours)} & $\mathbf{7.2}$ & $\mathbf{6.5}$ & $\mathbf{5.6}$ & $\mathbf{0\text{h}\,23\text{min}}$ \\
      \bottomrule
    \end{tabular}

  \end{table}
  We profile one steady-state TEAL round, using round 2 on CamVid with local budget 20, 1px budget, batch size 4, and a single RTX 6000 PRO GPU. CUDA memory is measured with \texttt{torch.cuda.max\_memory\_allocated()}, resetting peak statistics at every phase boundary. Per-image local herding over the $16\times16$ DINOv3 token grid is lightweight, peaking at only 0.36 GB of CUDA memory across 367 calls. The single cross-image global MaxHerding sweep over the union of token candidates is the main acquisition memory bottleneck, peaking at 1.14 GB. For the fully supervised performance, training the Mini-DPT decoder on four frozen DINOv3 layers for 50 epochs peaks at 0.47 GB, indicating that the decoder itself is not memory-intensive. Overall, a full round including Stage-2 uncertainty, training, and validation fits comfortably below 1.2 GB of VRAM.

\section{Implementation details}

\subsection{Mini-DPT head: changes from the Original Mini-DPT}
\label{app:minidpt}

Our variant of Mini-DPT head follows the original Mini-DPT decoder~\citep{yang2025segdino} with
two changes (Table~\ref{tab:minidpt-changes}). First, instead of bilinear
resizing followed by a $3\times3$ convolution, each branch is resampled with
the original DPT \citep{ranftl2021vision} reassemble mechanism: learned transposed convolutions for
upsampling, an identity at $\times1$, and a strided convolution for
downsampling. Second, the feature pyramid is one octave coarser, giving a
$64\times64$ decoder grid instead of $128\times128$; this grid also defines
the cells scored in Stage~2. Neither change hurts fully supervised accuracy:
with a DINOv3 ViT-B/16 backbone, Mini-DPT reaches $62.1$ mIoU on CamVid and
$43.0$ on ADE20K, compared with $61.7$ and $42.6$ for the original Mini-DPT. Other
implementation differences (initialization, a bypassed LayerNorm, and
\texttt{align\_corners} in the final resize) do not change the decoder
design.

\begin{table}[H]
\centering
\caption{\textbf{Architectural differences between the original Mini-DPT decoder and
our Mini-DPT head.} Grid sizes are for a $256\times256$ input.}
\label{tab:minidpt-changes}

\vspace{0.4em}

\scriptsize
\setlength{\tabcolsep}{6pt}
\renewcommand{\arraystretch}{1.15}
\begin{tabular}{@{}p{0.2\linewidth} p{0.33\linewidth} p{0.37\linewidth}@{}}
\toprule
\textbf{Component} & \textbf{Original Mini-DPT} & \textbf{Mini-DPT (ours)} \\
\midrule
Branch resampling
  & Bilinear resize, then learned $3\times3$ conv
  & Transposed conv (kernel $=$ stride) for $\times4$, $\times2$;
    identity for $\times1$; strided $3\times3$ conv for $\times\tfrac{1}{2}$ \\
Resampling parameters
  & 589,824
  & 475,136 \\
Pyramid scales
  & $[8, 4, 2, 1]$
  & $[4, 2, 1, \tfrac{1}{2}]$ \\
Decoder grid
  & $128\times128$ (2-px cells)
  & $64\times64$ (4-px cells) \\
\bottomrule
\end{tabular}
\end{table}
\subsection{Algorithm}
  \begin{algorithm}[H]
  \caption{TEAL --- Token Space Efficient Active
Learning at Round $t$}
  \label{alg:teal}
  \begin{algorithmic}[1]
  \REQUIRE Image set $\mathcal{D}=\{X_n\}_{n=1}^{N}$,
           labeled pixel set $\mathcal{L}_t$,
           frozen DINOv3 token features $\{\mathbf{h}_\tau\}$,
           Mini-DPT segmentation head $s_\theta$ trained on $\mathcal{L}_t$,
           per-image token budget $k$,
           per-round annotation budget $m$.
  \ENSURE Updated labeled pixel set $\mathcal{L}_{t+1}$.

  \STATE $\mathcal{M}_t \gets \emptyset$
         \hfill\COMMENT{aggregated per-image candidates}
  \STATE \emph{// Stage 1: Diversity selection in token space}
  \FOR{each image $X_n \in \mathcal{D}$}
      \STATE Let $\mathcal{U}_{n,t}$ be the valid unlabeled tokens of $X_n$.
      \STATE $\mathcal{M}_t \gets \mathcal{M}_t \cup MaxHerd(\mathcal{U}_{n,t},\,size = k)$
            \hfill\COMMENT{$k$ representative tokens per image}
  \ENDFOR
  \STATE $\mathcal{C}_t \gets \mathrm{MaxHerd}(\mathcal{M}_t,\,size=|\mathcal{M}_t|/2)$
         \hfill\COMMENT{global candidate token set}

  \STATE \emph{// Stage 2: Uncertainty selection on the decoder grid}
  \STATE Determine the set of valid decoder cells from the spatial support of $\mathcal{C}_t$.
  \FOR{each cell $v \in \{\textit{valid cells}\}$}
      \STATE $u(v) \gets p_v^{(1)} - p_v^{(2)}$
             \hfill\COMMENT{margin score from $s_\theta$}
  \ENDFOR
  \STATE $\mathcal{V}_t \gets$ the top $m$ valid cells with smallest $u(v)$
  \STATE $\mathcal{L}_{t+1} \gets \mathcal{L}_t \cup
         \bigl\{\mathrm{centre\_pixel}(v) : v \in \mathcal{V}_t \bigr\}$
  \RETURN $\mathcal{L}_{t+1}$
  \end{algorithmic}
  \end{algorithm}

\section{Qualitative Results}

\begin{figure}[H]
  \centering
  \scriptsize
  \setlength{\tabcolsep}{2pt}
  \renewcommand{\arraystretch}{1.0}

  \newcommand{\imgcell}[1]{%
    \begin{minipage}[c]{0.155\textwidth}\centering
      \includegraphics[width=\linewidth]{#1}
    \end{minipage}}

  \newcommand{\todocell}{%
    \begin{minipage}[c]{0.155\textwidth}\centering
      \fbox{\parbox[c][0.12\textwidth][c]{0.85\linewidth}{\centering
        \scriptsize\color{gray}\emph{TBA}}}
    \end{minipage}}

  \begin{tabular}{c c c c c c}
    \toprule
    \textbf{Dataset} & \textbf{Input} & \textbf{GT} & \textbf{eDALD (DDPM)} & \textbf{eDALD (DINOv3)} & \textbf{\method{} (Ours)} \\
    \midrule
    \rotatebox[origin=c]{90}{CamVid}
      & \imgcell{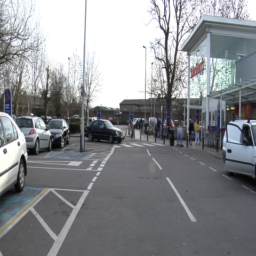}
      & \imgcell{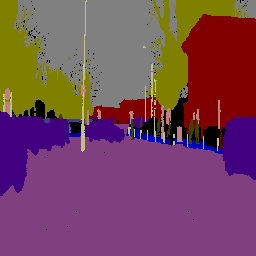}
      & \imgcell{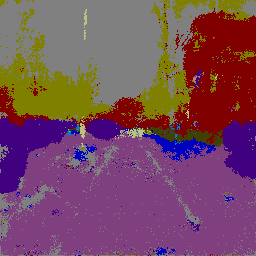}
      & \imgcell{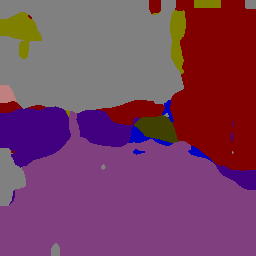}
      & \imgcell{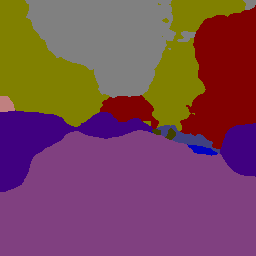} \\
    \midrule
    \rotatebox[origin=c]{90}{ADE20K}
      & \imgcell{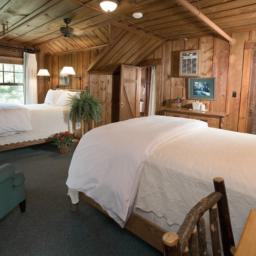}
      & \imgcell{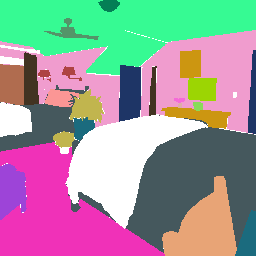}
      & \imgcell{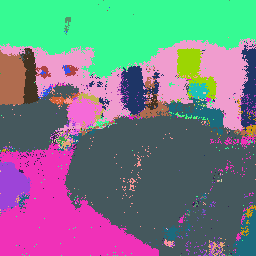}
      & \imgcell{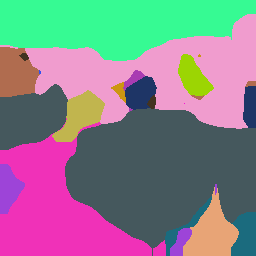}
      & \imgcell{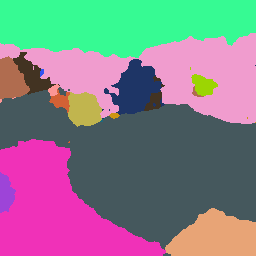} \\
    \midrule
    \rotatebox[origin=c]{90}{Cityscapes}
      & \imgcell{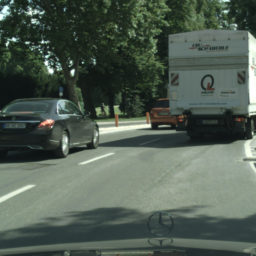}
      & \imgcell{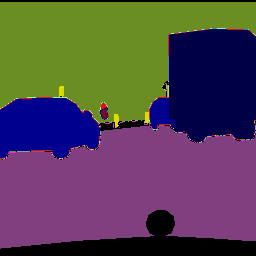}
      & \imgcell{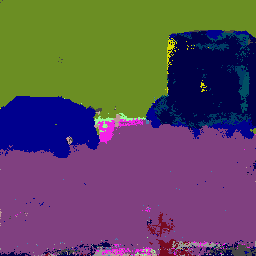}
      & \imgcell{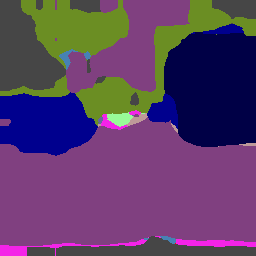}
      & \imgcell{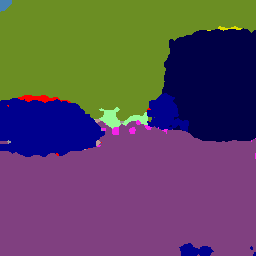} \\
    \bottomrule
  \end{tabular}

  \caption{
    \textbf{Qualitative segmentation maps at the final active-learning round.}
    We compare \method{} against eDALD with DDPM features and a matched-backbone eDALD using DINOv3 features.
    \method{} produces cleaner object regions and fewer boundary artifacts, consistent with the quantitative gains in the low-budget setting.
  }
  \label{fig:qualitative-best}
\end{figure}

\section{Broader Impact}
Our work has a positive societal impact by significantly reducing the annotation cost required to train dense prediction models. This is especially important in domains where reliable annotations are difficult to obtain, such as disaster response, rare disease diagnosis, and manufacturing inspection. This could make segmentation models more accessible to teams with limited labeling resources, including smaller research groups and organizations working in resource-constrained settings.

At the same time, caution should be taken when applying TEAL to high-stakes or sensitive domains, such as law enforcement, medical decision-making, or applications involving private information. Because TEAL relies on a small amount of supervision, there is an increased risk that the model may replicate biases or vulnerabilities present in the training data. Thus, TEAL should be used only after testing and validating carefully on realistic data, checking whether it works fairly across different groups, and keeping humans involved in important decisions.

\newpage

\end{document}